\documentclass[conference]{IEEEtran}
\pdfoutput=1
\IEEEoverridecommandlockouts                              

\usepackage{cite}
\usepackage{amsmath,amssymb,amsfonts}
\usepackage{graphicx}
\usepackage{textcomp}
\usepackage{xcolor}
\usepackage{booktabs}
\usepackage{multirow}
\usepackage{array}
\usepackage{tabularx}
\usepackage{colortbl}
\usepackage[caption=false,font=footnotesize]{subfig}

\usepackage{hyperref}

\newcommand{\OUT}[1]{}

\title{\LARGE \bf Learning to Plan in Human-Robot Collaboration: Multimodal Reinforcement Learning for Adaptive Interaction}

\author{Afagh Mehri Shervedani$^{1}$, Siyu Li$^{1}$, Natawut Monaikul$^{1}$, Bahareh Abbasi$^{2}$, Barbara Di Eugenio$^{1}$, \\and Milo\v s \v Zefran$^{1}$
\thanks{$^{1}$A. Mehri Shervedani, S. Li, N. Monaikul, B. Di Eugenio, and M. \v Zefran are with the University of Illinois Chicago, Chicago, IL 60607 USA.}%
\thanks{$^{2}$B. Abbasi is with California State University Channel Islands}%
\thanks{This work has been supported by the National Science Foundation grants IIS-1705058, CMMI-1762924, CCF-2240532, and ECCS-2217023.}%

\thanks{This study has been published in the International Journal of Social Robotics\cite{mehri2025multimodal}.}}

\begin{document}

\maketitle

\begin{abstract}
Robot assistants for older adults and people with disabilities need to perform collaborative tasks with users effectively. The core component of these systems is an interaction manager whose job is to observe and assess the task and infer the state of the human and their intent for the robot to choose the best course of action. Due to the sparseness of the data in this domain, the policy for such multimodal systems is often crafted by hand; as the complexity of interactions grows, this process is not scalable. This paper proposes a reinforcement learning (RL) approach to automatically generate the multimodal policy of the robot. Our system focuses on a realistic scenario where a robot assists a user in locating objects within a home environment, managing multimodal signals, including language and physical actions, to select the best action. In contrast to traditional dialog systems, our agent is trained with a simulator that uses human data and can deal with multiple modalities. We use a simple high-level reward function that needs no fine-tuning and enforce some preconditions to speed up the training process. A human study evaluating the system in a real-world setting demonstrates promising results, indicating high usability and effective task completion. This RL-based approach offers a scalable and interpretable alternative for designing interaction managers in multimodal human-robot collaborations.
\end{abstract}

\section{Introduction}
\label{sec:Introduction}

\begin{figure}[t]
\centering
\includegraphics[width=\columnwidth]{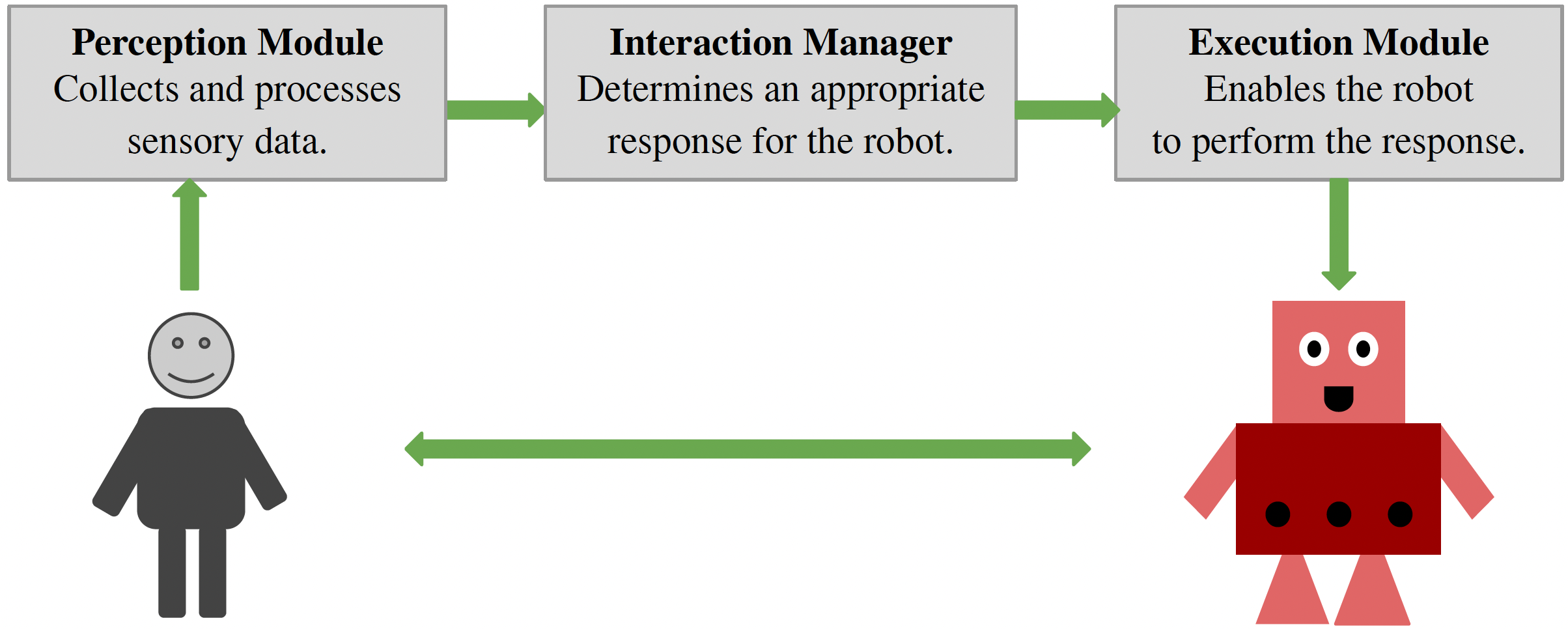}
\caption{The \emph{Sense-Plan-Act} cycle in an assistive robot}
\label{fig:sense-plan-act}
\end{figure}

Assistive robots that support older adults and people with disabilities in activities of daily living (ADLs) must collaborate with users through multimodal interactions involving speech, gestures, and physical actions. Such robots typically follow a \emph{sense-plan-act} cycle, in which perception interprets the environment and user behavior, an interaction manager selects the robot's response, and an execution module performs the selected action (Fig.~\ref{fig:sense-plan-act}).

We previously developed Hierarchical Bipartite Action-Transition Networks (HBATNs) as an interaction-management architecture for multimodal human-robot collaboration~\cite{8968505, monaikul2020role}, grounded in the \textbf{ELDERLY-AT-HOME} corpus of interactions between elderly individuals (ELD) and nursing students (Helper, HEL) during ADLs~\cite{chen2015roles}. We focused on the \emph{Find} task, in which the participants collaboratively locate a hidden object using speech, pointing gestures, and \emph{haptic-ostensive} (H-O) actions that bring physical objects into conversational focus~\cite{chen2015roles}. Although HBATNs provided effective interaction policies, their construction was performed manually and would need to be redesigned for each new task.

Large Language Models (LLMs) have recently enabled robots to leverage broad world knowledge and reasoning capabilities for complex policies~\cite{ouyang2022training, brown2020language, achiam2023gpt}. However, LLM-based approaches often require substantial data and remain challenging to ground in multimodal robotic interactions~\cite{brohan2023can, driess2023palm, wu2023mllm, chen2024scalable, Li_2024_CVPR, wu2024nextgpt, caffagni-etal-2024-revolution}. We therefore propose an interpretable reinforcement learning (RL)-based interaction manager that learns a collaborative policy through trial-and-error interaction with an environment~\cite{sutton2018reinforcement,sallans2004reinforcement}. We train the robot to assume the HEL role while collaborating with a human in the ELD role during the \emph{Find} task.

A key challenge is providing an interactive training environment that can realistically model human behavior. We address this challenge with a neural network-based user simulator trained on the \emph{Find} task data from ELDERLY-AT-HOME~\cite{10309444}. Inspired by Behavioral Cloning~\cite{bratko1995behavioural, torabi2018behavioral}, the simulator models human responses during interaction and is augmented with synthetic, data-driven misunderstandings to account for the limited corpus size and errors that arise in robotic interactions. An error-injection module further exposes the RL agent to imperfect interactions during training.

Our main contribution is an interpretable RL-based interaction manager for multimodal collaborative robots, together with an end-to-end user simulator that enables scalable training from limited human interaction data. We evaluate the resulting system through a human-subject user study and assess both the RL-based interaction manager and the perception and execution components. The results demonstrate high task accuracy and user satisfaction, showing the potential of RL for scalable interaction management in assistive robotics.

\section{Reinforcement Learning Framework}
\label{sec:Framework}

We formulate interaction management as a Deep-Q-Network (DQN)~\cite{mnih2013playing}, where the assistive robot (HEL) learns a policy by interacting with the neural network-based user simulator. Because learning appropriate behavior from a simple reward function alone is difficult with limited interaction data, we first warm up the agent using DA{\footnotesize GGER}, an Imitation Learning (IL) algorithm~\cite{ross2011reduction}. The resulting policy is then used to initialize the DQN policy and target networks.

\begin{figure}[t]
\centering
\includegraphics[width=\columnwidth]{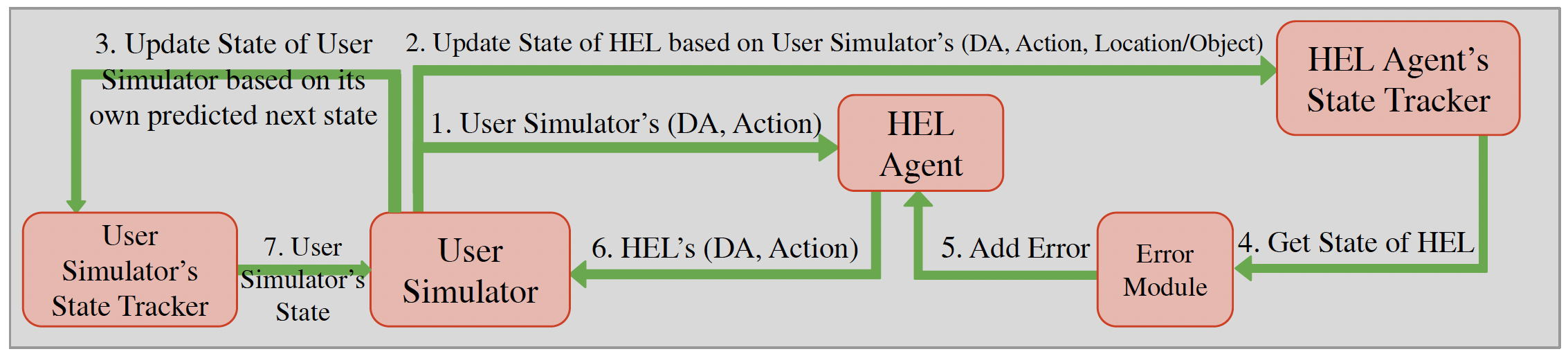}
\caption{The interaction between the User Simulator and the HEL agent during RL}
\label{fig:arch_IL_RL}
\end{figure}

As shown in Fig.~\ref{fig:arch_IL_RL}, the HEL agent receives the user simulator's action and its previous state and outputs a \emph{Dialogue Act} (DA) and physical action. The state is represented by three variables: target-object type ($O_T$), potential location ($L$), and target object ($O$). Each variable takes one of three values: UNKNOWN, MATCH, or MISMATCH, indicating whether the corresponding information is undetermined, agrees with the user's state, or differs from it.

\subsection{Model Architecture and Warm-up}
The HEL agent uses a lightweight fully connected network consisting of two fully connected layers, dropout ($0.1$), and an output layer with ReLU activation. We evaluated alternative architectures, including BERT, ALBERT, DistilBERT, and Transformer-XL, but found that the simpler network was more effective and computationally efficient for the relatively small interaction dataset.

\begin{figure}[h]
\centering
\subfloat[{\scriptsize Training Loss}\label{fig:dagger_loss}]{%
\includegraphics[width=0.47\columnwidth]{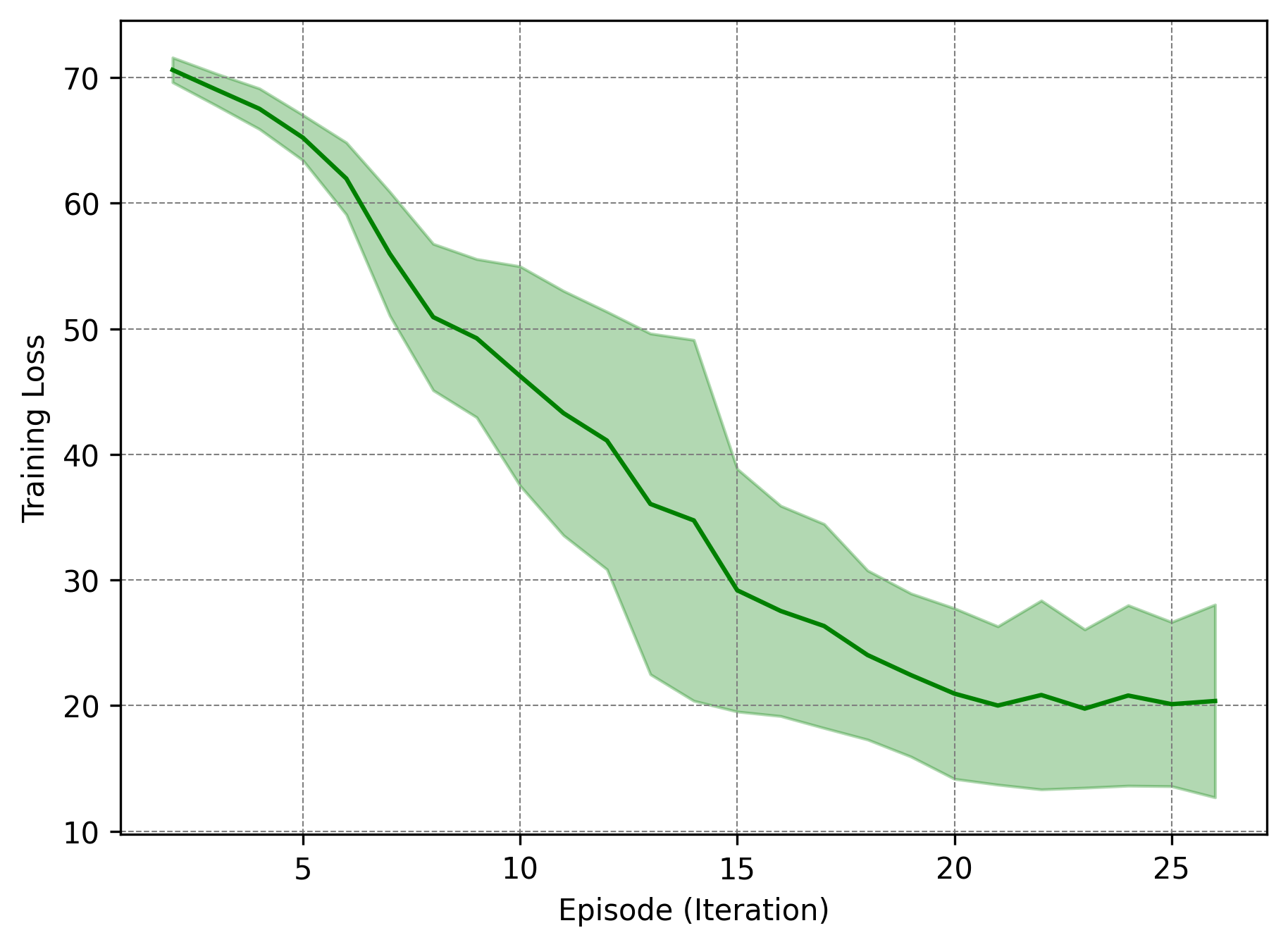}}
\hfill
\subfloat[{\scriptsize Success Rate and Average Turns}\label{fig:dagger_s_rate}]{%
\includegraphics[width=0.52\columnwidth]{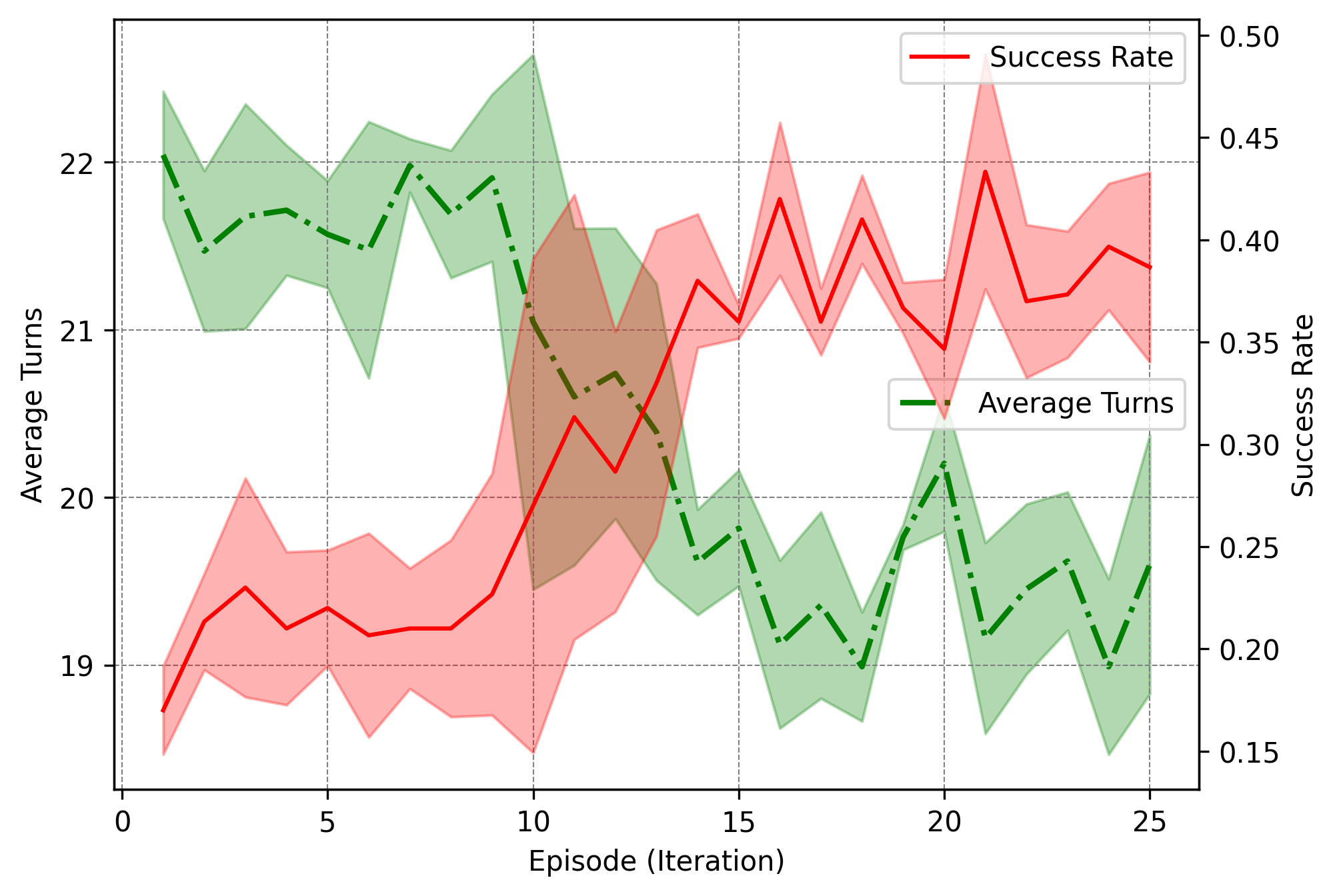}}
\caption{DA{\footnotesize GGER} Algorithm Training Evaluations}
\label{fig:dagger_evaluations}
\end{figure}

During DA{\footnotesize GGER} warm-up, the agent interacts with the user simulator while expert actions extracted from the ELDERLY-AT-HOME corpus are recorded but not executed. These expert corrections are iteratively aggregated to train the policy~\cite{ross2011reduction}. To expose the agent to mismatched beliefs that are relatively rare in the original corpus, an error module converts MATCH states to MISMATCH in 25\% of applicable cases. We run DA{\footnotesize GGER} for 25 episodes, with each episode containing at most 25 interaction turns. Rather than using the fully trained DA{\footnotesize GGER} policy, we initialize DQN using the policy obtained after 10 episodes, before overfitting to the limited demonstrations occurs (Fig.~\ref{fig:dagger_evaluations}).

\subsection{Deep-Q-Learning}
\begin{figure}[t]
\centering
\subfloat[{\scriptsize Training Loss and Average Reward}\label{fig:dql_loss}]{%
\includegraphics[width=0.503\columnwidth]{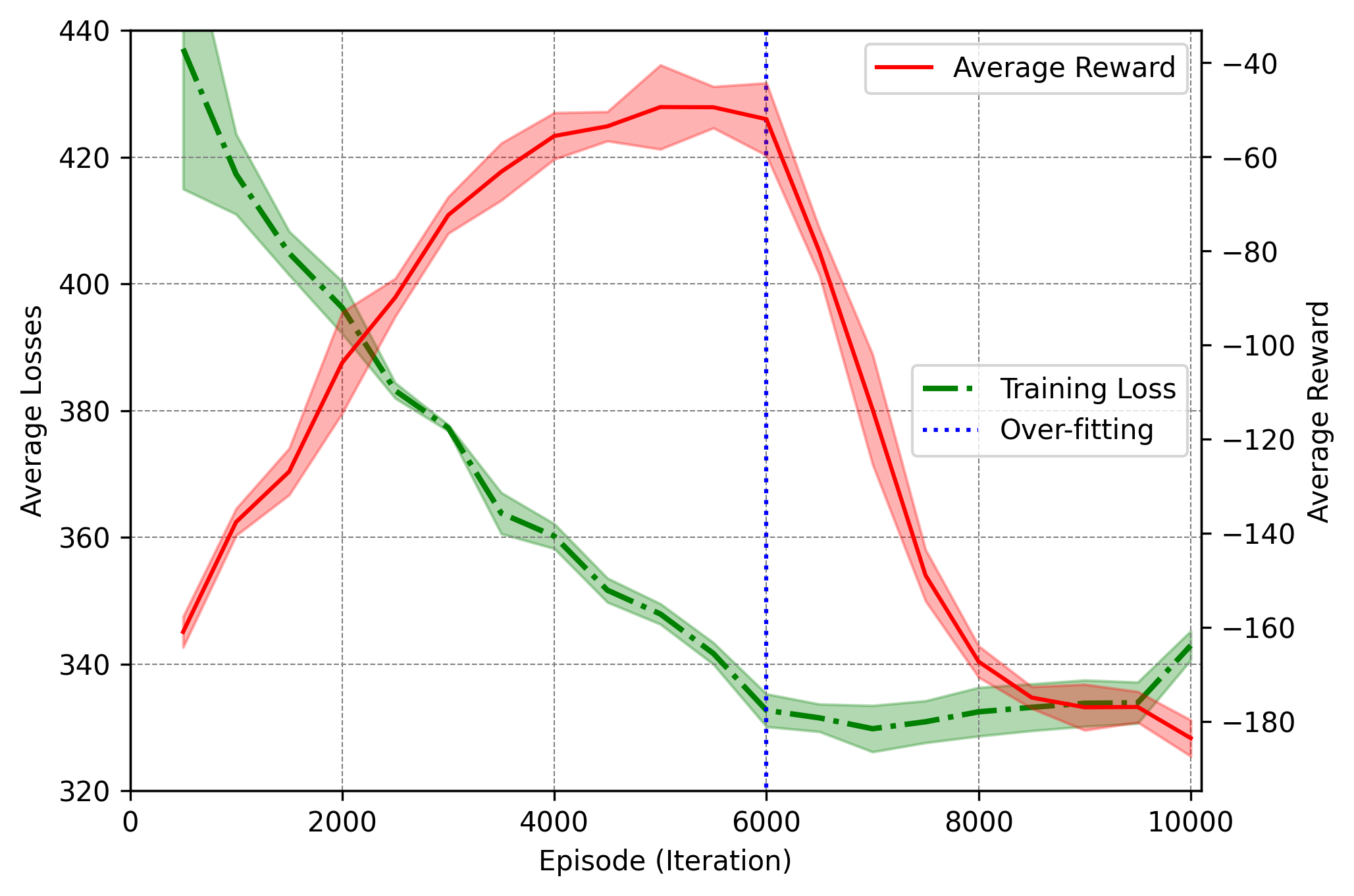}}
\hfill
\subfloat[{\scriptsize Success Rate and Average Turns}\label{fig:dql_s_rate}]{%
\includegraphics[width=0.49\columnwidth]{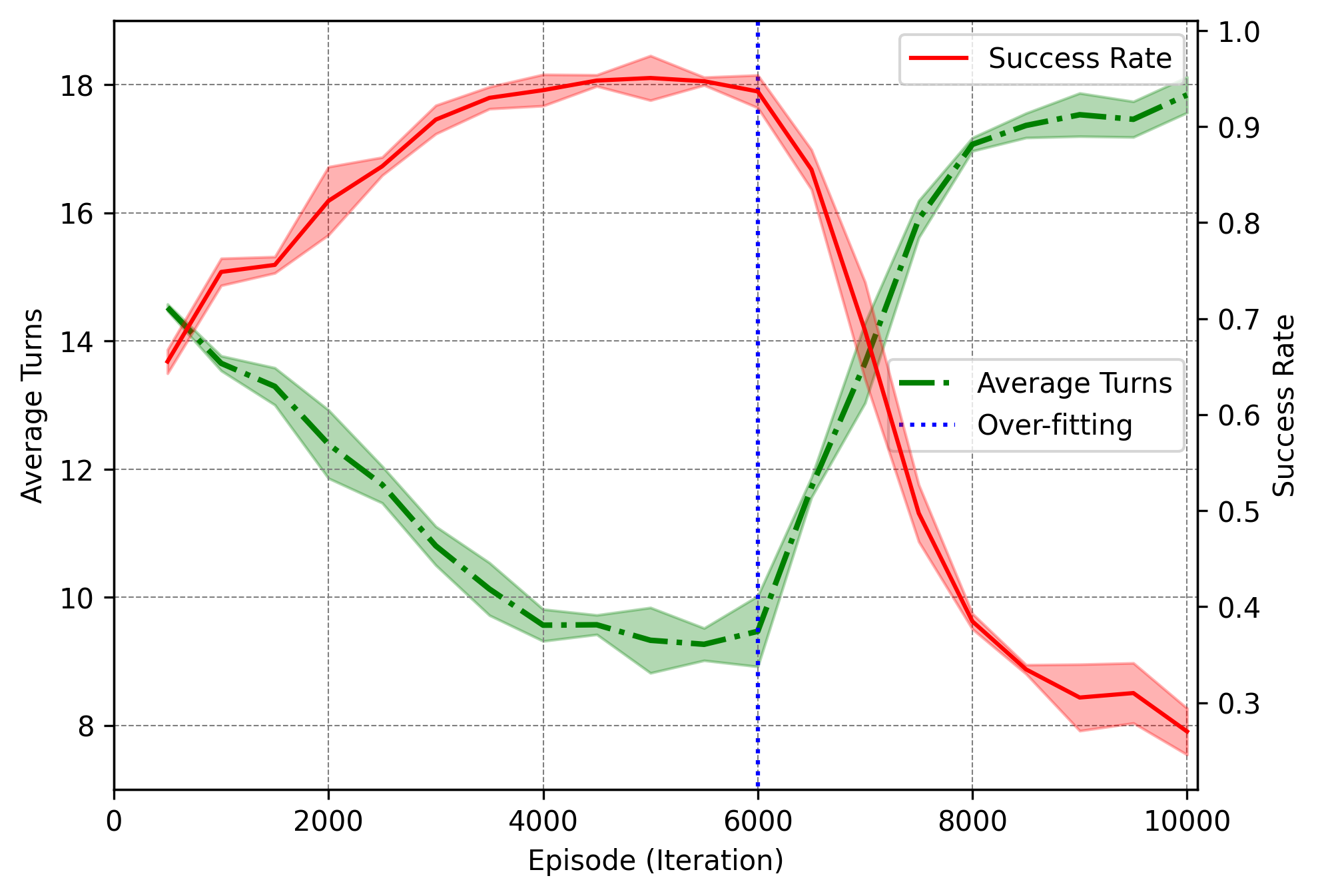}}
\caption{DQL Algorithm Training Evaluations}
\label{fig:dql_evaluations}
\end{figure}

\begin{figure}[t]
\centering
\subfloat[{\scriptsize Training Loss and Average Reward}\label{fig:dql_only_loss}]{%
\includegraphics[width=0.49\columnwidth]{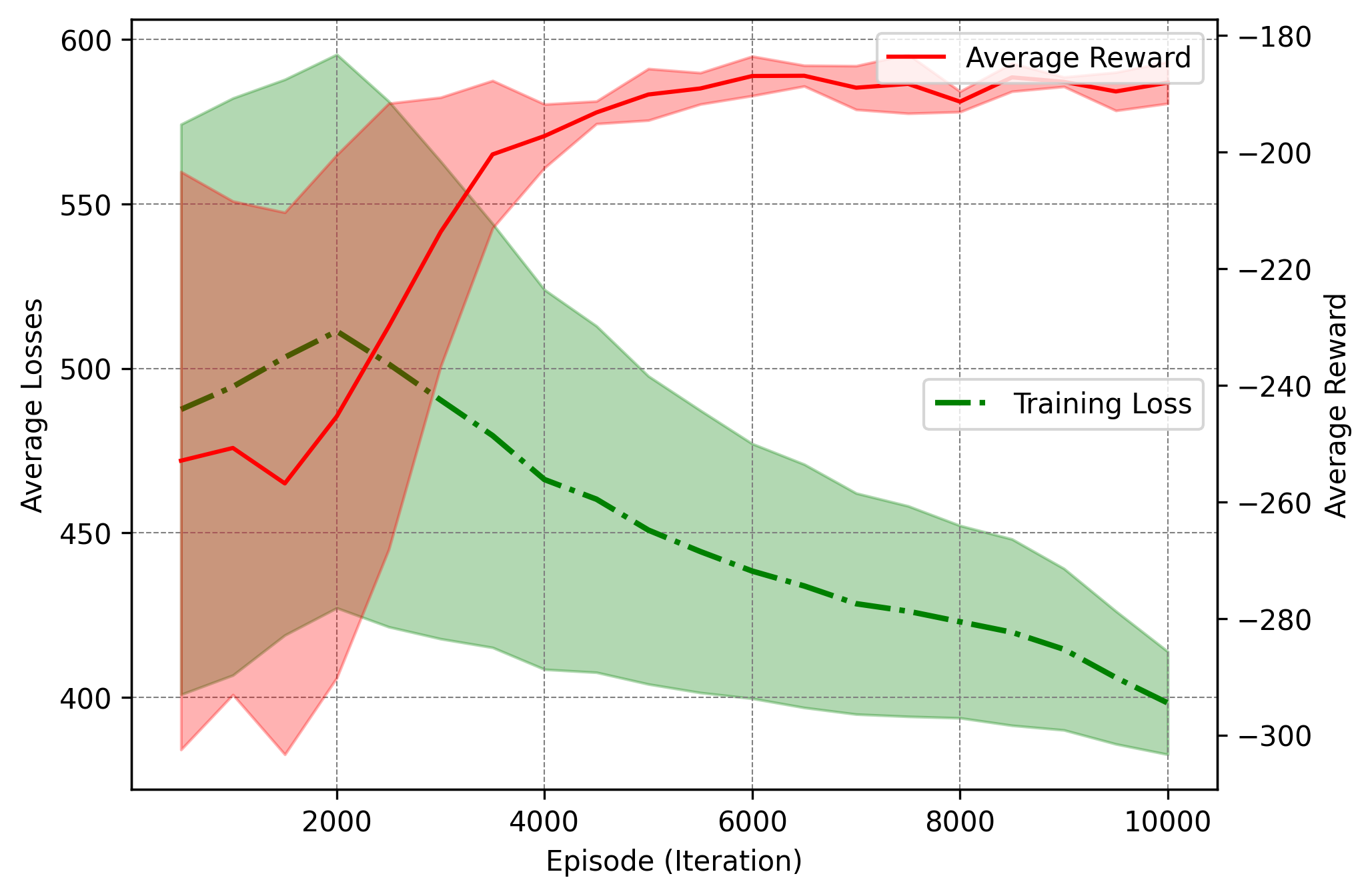}}
\hfill
\subfloat[{\scriptsize Success Rate and Average Turns}\label{fig:dql_only_s_rate}]{%
\includegraphics[width=0.495\columnwidth]{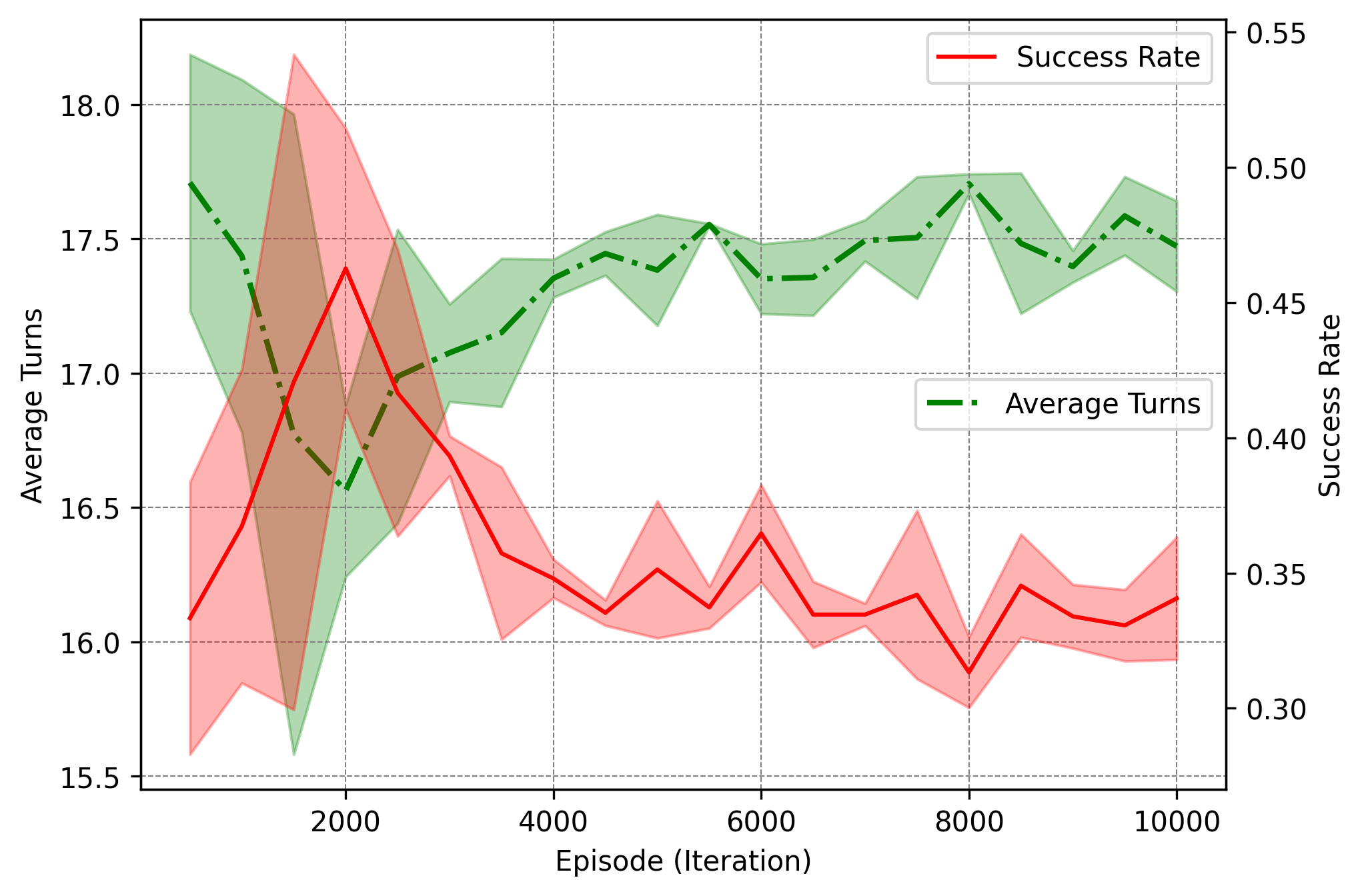}}
\caption{DQL Without Warm-up}
\label{fig:dql_only_evaluations}
\end{figure}
The DQN policy is trained using a policy network and a target network initialized with the DA{\footnotesize GGER} weights. This warm-up provides a reasonable initial policy and reduces the need to encode complex human behavior in the reward function.

The reward function assigns a small penalty ($-r$, $r=1$) to each action to encourage efficient task completion, a positive reward ($+2r$) for successful completion, and a penalty ($-2r$) when the maximum number of turns ($M=20$) is reached unsuccessfully. Actions that violate task preconditions receive a large penalty ($-Z$, $Z=50$). Specifically, the agent cannot verify $O_T$ before it is provided by the user, verify $L$ before it is provided, or verify $O$ before both $O_T$ and $L$ are available. The resulting reward function is:


\begin{equation}
R(\mathrm{success},\mathrm{reached\text{-}M},\mathrm{illegal}) =
\begin{cases}
-Z, & \text{if illegal} = \text{True},\\
+2r, & \text{if success} = \text{True},\\
-2r, & \substack{\text{if reached-}M = \text{True and}\\
\text{success} = \text{False}},\\
-r, & \text{otherwise}.
\end{cases}
\label{eq:reward-function}
\end{equation}

During training, transitions $(s,r,a,s')$ are stored in replay memory. The policy network is optimized every $C=500$ episodes, and its weights are transferred to the target network every $mC=2000$ episodes. Training is stopped before the degradation observed after approximately 6000 episodes, when the replay memory begins replacing older samples and performance decreases (Fig.~\ref{fig:dql_evaluations}).

Figure~\ref{fig:dql_only_evaluations} further demonstrates the importance of the DA{\footnotesize GGER} warm-up. Compared with training DQN from scratch, the warmed-up agent converges faster, achieves higher rewards and success rates, and stabilizes earlier. Even after 10,000 episodes, the agent trained without warm-up does not reach the initial performance of the DA{\footnotesize GGER}-initialized agent.

\section{Experimental Evaluations}
\label{sec:Evaluations}

\begin{table*}[t]
\caption{Summary of user-study results}\label{table:quant_results}
\begin{tabular*}{\textwidth}{@{\extracolsep\fill}lcccccc@{}}
\toprule
\textbf{Avg.} & \textbf{Success}& \textbf{Overall System} & \textbf{SSREs} & \textbf{Wrong} & \textbf{Wrong} & \textbf{RL Policy} \\
\textbf{\#Turns} & \textbf{Rate} & \textbf{Accuracy} &  & \textbf{DAs} &  \textbf{Pointing} &\textbf{(IM) Acc.} \\
\midrule
6.3 & 96\% & 90.2\% & 6.4\% & 15.1\% & 13.6\% & $\approx$97.8\% \\
\bottomrule
\end{tabular*}
\end{table*}

\setlength{\tabcolsep}{1pt}
\begin{table*}[t]
\caption{RL-based and HBATN-based systems comparison}\label{table:systems-comparison}
\footnotesize
\begin{tabular*}{\textwidth}{@{\extracolsep\fill}l|c|c|c|c|c|c|c|c@{}}
\toprule
System& Avg.  &Success&System &SSREs&Wrong&Wrong    &IM &IM Contribution to\\
      &\#Moves&Rate   &Acc. &     &DAs  &Pointings& Acc.&Non-elig. Responses\\
\midrule
RL-based & 12.6 & 96.0\% & 90.2\% & 6.4\% & 15.1\% & 13.6\% & 97.8\% & 15.2\%\\
HBATN-based & 15.6 & 85.7\% & 89.8\% & 17.3\% & 17.7\% & 28.9\% & 96.9\% & 29.2\%\\
\bottomrule
\end{tabular*}
\end{table*}

\subsection{Robot Implementation}
\label{subsection:Implementation}

\begin{figure}[t]
\centering
\includegraphics[width=\columnwidth]{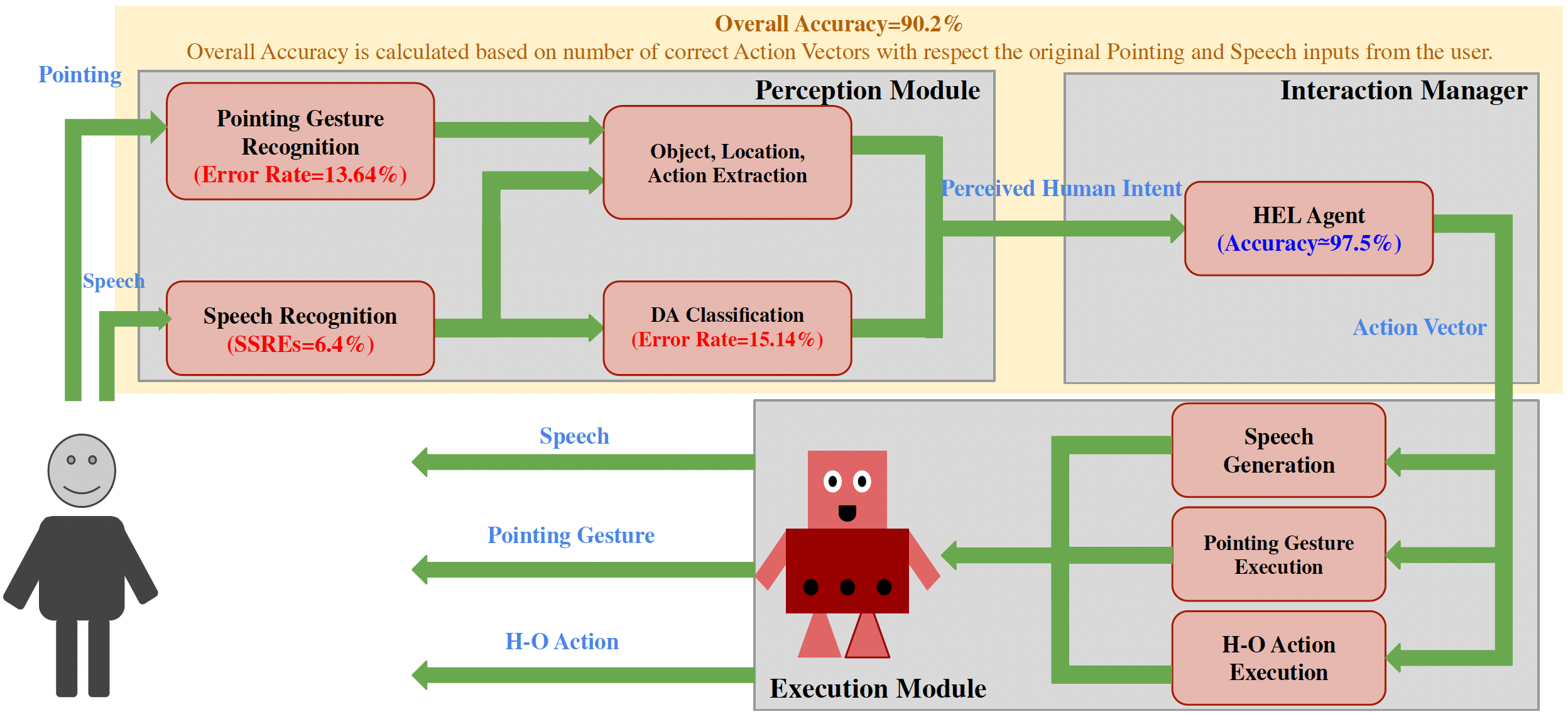}
\caption{The \emph{Sense-Plan-Act} cycle in an assistive robot and the performance of different system components}
\label{fig:sense-plan-act-detail}
\end{figure}

We implemented the proposed framework on a Baxter robot from Rethink Robotics to enable multimodal interaction with human users. The implementation follows the perception and execution modules shown in Fig.~\ref{fig:sense-plan-act-detail}. The perception module converts human speech and gestures into the Perceived Human Intent representation required by the RL-based Interaction Manager. Speech is transcribed using the Google Cloud Speech-to-Text API~\cite{s2t_api}, and an ALBERT-based dialogue-act (DA) classifier~\cite{lan2019albert} predicts the user's DA from the transcribed utterance. The classifier was trained on 7,617 ELDERLY-AT-HOME datapoints (80/10/10 train/validation/test split) with data augmentation and achieved 88.41\% test accuracy. An action extractor identifies target objects and locations from speech and pointing gestures using an NLTK-based dictionary~\cite{bird2009natural}, while pointing gestures are detected using the model from~\cite{8968505}.

The execution module converts the RL agent's action vector into speech and physical actions. A rule-based text generator converts DA tags into utterances, which are synthesized using Pyttsx3~\cite{pyttsx3}. Baxter executes physical actions, including pointing, drawer manipulation, and visually presenting objects, in parallel with speech generation.

\subsection{User Study}
\begin{figure}[t]
\centering
\includegraphics[width=\columnwidth]{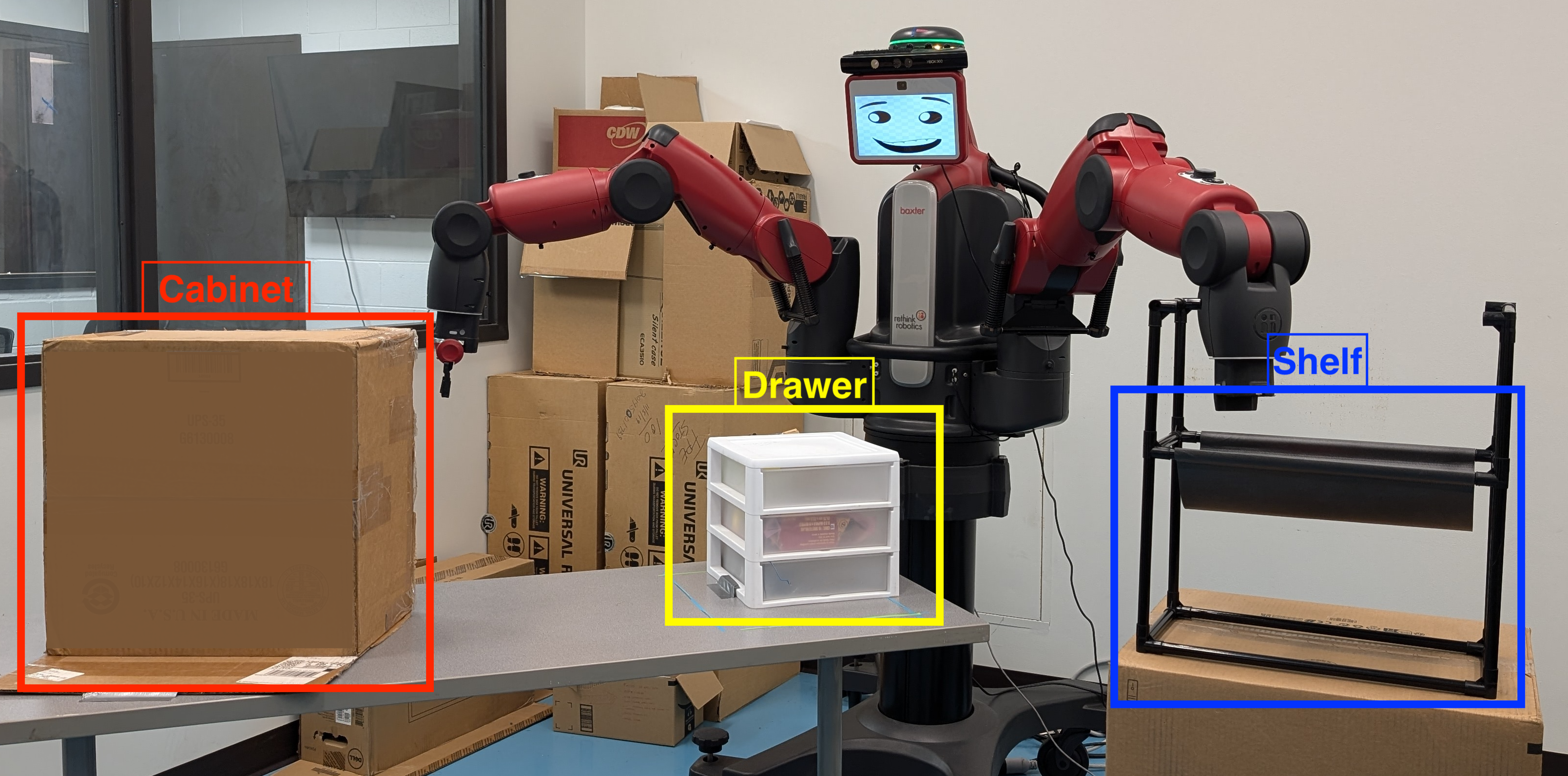}
\caption{The experimental setup}
\label{fig:exp_env}
\end{figure}
We conducted a user study with 12 healthy adults (6 women, 6 men), each completing 5--7 trials, for a total of 75 trials. The experimental environment contained a drawer, shelf, and cabinet with colored cups and balls placed randomly across locations (Fig.~\ref{fig:exp_env}). In each trial, the participant selected a target object without knowing its location and collaborated with the robot to find it.

\subsection{Results}
\label{subsection:quant-results}

The system achieved a 96.0\% task success rate with an average of 6.3 turns per interaction. Across 469 human inputs, 46 robot responses were non-eligible, resulting in 90.2\% overall system accuracy. Component-level error rates were 6.4\% for serious speech recognition errors (SSREs), 15.1\% for incorrect DAs, and 13.6\% for incorrect pointing gestures. When the input was correctly perceived, the RL-based Interaction Manager achieved approximately 97.8\% accuracy. Only 15.2\% of the system's non-eligible responses were attributable to the RL policy itself, indicating that most errors originated in perception or human actions.


To examine whether repeated trials introduced a learning effect, we compared the first three trials of each subject (Set 1) with the remaining trials (Set 2) using Average Turns, Success Rate, and System Accuracy. Based on Shapiro--Wilk tests, we used a paired $t$-test for Average Turns and Wilcoxon signed-rank tests for Success Rate and System Accuracy. The results showed no significant difference in Average Turns ($p=0.378$) or Success Rate ($p=0.655$), while System Accuracy significantly increased from 86.6\% to 95.1\% ($p=0.036$). Thus, repeated trials may have influenced system accuracy, although task-level efficiency and success were not significantly affected.

\subsection{Comparison with HBATNs}
\label{subsec:Comparison}

We compared the proposed RL-based system with the HBATN-based system from~\cite{8968505}. The RL-based system reduced average interaction length from 15.6 to 12.6 moves and increased success rate from 85.7\% to 96.0\%. It also reduced SSREs from 17.3\% to 6.4\% and incorrect pointing gestures from 28.9\% to 13.6\%, while DA classification and Interaction Manager accuracy remained comparable. The reduction in SSREs was statistically significant ($p=0.008$), as were the improvements in average interaction length ($p=0.017$) and success rate ($p=0.010$). The RL policy achieved 97.8\% accuracy compared with 96.9\% for HBATNs, while accounting for a smaller proportion of non-eligible responses (15.2\% vs.~29.2\%). These results indicate that the automatically learned RL policy performs comparably to the manually constructed HBATNs policy while providing a more scalable approach to interaction management.

\subsection{Survey}
\label{subsection:survey-results}

\section{Discussion}
\label{sec:Discussion}

Following the experiment, participants completed a 15-question survey designed according to established guidelines~\cite{hoffman2020primer, schrumConcerningTrendsLikert2023}. The questions were grouped into five categories: Responsiveness, Comfort, Familiarity and Learning, Fatigue and Habituation, and Overall Quality. Cronbach's alpha was calculated to assess the internal consistency of the multi-item groups. As shown in Table~\ref{table:survey_results}, participants reported high Responsiveness (6.03/7) and Comfort (5.73/7), as well as high Familiarity and Learning (4.33/5). Fatigue and Habituation received a low score (1.69/5), indicating limited fatigue, frustration, or boredom. The Overall Quality score was 4.25/5. The Cronbach's alpha values for the multi-item groups (0.71--0.82) indicate acceptable to good internal consistency.

The Familiarity and Learning results suggest that participants became more comfortable interacting with the robot over repeated trials. However, the learning-effect analysis in Section~\ref{subsection:quant-results} showed no significant changes in Average Turns or Success Rate, although System Accuracy differed significantly between the first and later trials. Thus, participants may have adapted their interaction strategies without substantially affecting task-level performance. The low Fatigue and Habituation score further suggests that repeated trials did not substantially reduce engagement.

We can directly compare only the Overall Quality score with the HBATN-based system~\cite{8968505}, because that study reported only this survey measure and did not provide subject-level responses. The HBATN-based system received an average score of 4/5, compared with 4.25/5 for our system. A Mann--Whitney U test yielded $p=0.32$, indicating no statistically significant difference in perceived overall interaction quality.

\begin{table}[t]
\caption{Statistics of responses to the survey}\label{table:survey_results}
\begin{tabular*}{\columnwidth}{@{\extracolsep{\fill}}llllll@{}}
\toprule
\scriptsize{Question \#} & G1 & G2 & G3 & G4 & G5\\
\midrule
\scriptsize{Average} & 6.03 & 5.73 & 4.33 & 1.69 & 4.25\\
\midrule
\scriptsize{Standard Deviation} & 0.81 & 0.8 & 0.77 & 0.72 & NA\\
\midrule
\scriptsize{Cronbach's alpha} & 0.82 & 0.71 & 0.75 & 0.82 & NA\\
\bottomrule
\end{tabular*}
\end{table}

\subsection{Limitations, Applications, and Future Work}
\label{subsection:Analysis}

Our comparison with the HBATN-based system has several limitations. Our study involved 12 participants, whereas the HBATN-based study involved 7, and subject-level data from the latter were unavailable. Consequently, for some metrics, we evaluated HBATN components using inputs collected in our study, while comparisons involving Average Turns, Success Rate, Wrong Pointings, and Overall Quality relied on aggregate values reported in~\cite{8968505}. Because subject-level HBATN data were unavailable, we assumed a uniform distribution of reported aggregate values when conducting statistical comparisons. This assumption may reduce the precision of the statistical analysis; therefore, the observed improvements should be interpreted with caution. Access to subject-level data from the HBATN study would enable a more rigorous comparison.

Despite these limitations, the results demonstrate the potential of the RL-based framework for multimodal human--robot interaction, particularly in socially assistive applications where robots must interpret both verbal and non-verbal user actions. The framework could be applied to healthcare assistance, education, and other settings requiring adaptive interaction.

A further limitation is task generalization. Although the RL framework can be applied to different tasks, each new task requires an appropriate user simulator, which in turn requires multimodal demonstration data and substantial development effort. Future work should investigate more data-efficient approaches for constructing task-specific simulators, as well as generic simulators capable of supporting multiple HRI tasks. Combining Behavioral Cloning with LLMs is a promising direction: LLMs can provide general language knowledge, while task-specific demonstrations can ground the simulator in the behaviors required for a particular HRI task.

Finally, we plan to extend the framework to physical Human--Robot Interaction (pHRI) tasks~\cite{Rysbek2024Proactive}. Such tasks require robots to reason about both human intent and physical interaction. Incorporating additional modalities, including force and torque sensing, could enable the robot to respond to physical cues and changes in user intent. Transfer learning could further facilitate adaptation of the learned RL policy to new pHRI tasks while reducing the amount of task-specific training required.

\section{Conclusion}
\label{sec:Conclusion}
This paper proposes a reinforcement learning (RL) framework for HRI, and in particular for the development of interaction managers for multimodal collaborative tasks involving older adults and individuals with disabilities. The approach overcomes the limitations of handcrafted policies and can deal with data sparsity, which is challenging for traditional methods. Our RL-based interaction manager, trained with a simulator using human data, demonstrates promising results in multimodal interactions, effectively handling speech, gestures, and physical actions. Through a comprehensive evaluation, we highlight the advantages of our RL-based system over the HBATN-based approach, particularly in terms of speech recognition accuracy, DA classification performance, and overall system quality. Notably, our system achieves a high level of user satisfaction, as evidenced by positive feedback from a human study. By leveraging RL, we provide a scalable solution for building interaction managers that can adapt to various tasks and environments. Future work may focus on further refining our RL framework, exploring additional applications, and generalizing it to other tasks. Overall, this research contributes to advancing the state-of-the-art in assistive robotics and lays the foundation for more intelligent and responsive robotic assistants in the future.

\bibliographystyle{IEEEtran}
\bibliography{references}

\appendices
\section{The survey questions}
\label{sec:Appendix_survey}
The survey questions are grouped as follows.
\bigskip

\begin{enumerate}
    
    \item \textbf{Responsiveness:}
    \begin{itemize}
        \item How confident were you that the robot would be able to help you find the object you were looking for? (7-point Likert, 1:Very Unconfident, 2:Unconfident, 3:Slightly Unconfident, 4:Neutral, 5:Slightly Confident, 6:Confident, 7:Very Confident)
        \item How well were you able to predict how the robot would act during the interaction? (7-point Likert, 1:Very Poor, 2:Poor, 3:Slightly Poor, 4:Neutral, 5:Well, 6:Slightly Well, 7:Very Well)
        \item How well was the robot able to respond to your actions during the interaction? (7-point Likert, 1:Very Poor, 2:Poor, 3:Slightly Poor, 4:Neutral, 5:Well, 6:Slightly Well, 7:Very Well)
        \item How well was the robot able to express what it was trying to do? (7-point Likert, 1:Very Poor, 2:Poor, 3:Slightly Poor, 4:Neutral, 5:Well, 6:Slightly Well, 7:Very Well)
        \item How well was the robot able to take the initiative and move the interaction forward when the instructions were not clear? (7-point Likert, 1:Very Poor, 2:Poor, 3:Slightly Poor, 4:Neutral, 5:Well, 6:Slightly Well, 7:Very Well)
    \end{itemize}
    \bigskip

    \item \textbf{Comfort:}
    \begin{itemize}
        \item How comfortable were you during the interaction? (7-point Likert, 1:Very Uncomfortable, 2:Uncomfortable, 3:Slightly Uncomfortable, 4:Neutral, 5:Slightly Comfortable, 6:Comfortable, 7:Very Comfortable)
        \item If you were to have an assistant to help you find objects from various locations, would it be acceptable to have this robot as your assistant? (under the condition that the places are reachable for the robot) (7-point Likert, 1:Very Unacceptable, 2:Unacceptable, 3:Slightly Unacceptable, 4:Neutral, 5:Slightly Acceptable, 6:Acceptable, 7:Very Acceptable)
        \item How similar was the robot’s behavior to how a human assistant would behave? (7-point Likert, 1:Very Unsimilar, 2:Unsimilar, 3:Slightly Unsimilar, 4:Neutral, 5:Slightly Similar, 6:Similar, 7:Very Similar)
        \item How well was the robot able to make the interaction natural and relaxed? (7-point Likert, 1:Very Poor, 2:Poor, 3:Slightly Poor, 4:Neutral, 5:Well, 6:Slightly Well, 7:Very Well)
    \end{itemize}
    \bigskip

    \item \textbf{Familiarity, Learning:}
    \begin{itemize}
        \item As you progressed through the trials, did you find it easier to interact with the robot? (5-point Likert scale, 1:Strongly Disagree, 2:Disagree, 3:Neutral, 4:Agree, 5:Strongly Agree)
        \item As you progressed through the trials, did you adapt your behavior? (5-point Likert scale, 1:Strongly Disagree, 2:Disagree, 3:Neutral, 4:Agree, 5:Strongly Agree)
    \end{itemize}
    \bigskip

    \item \textbf{Fatigue, Habituation:}
    \begin{itemize}
        \item How tired did you get as trials went on? (5-point Likert scale, 1:Not tired at all, 2:Slightly tired, 3:Moderately tired, 4:Tired, 5:Significantly tired)
        \item How frustrated did you get as trials went on? (5-point Likert scale, 1: Not frustrated at all, 2:Slightly frustrated, 3:Moderately frustrated, 4:Frustrated, 5: Significantly frustrated)
        \item How bored did you get as trials went on? (5-point Likert scale, 1: Not bored at all, 2:Slightly bored, 3:Moderately bored, 4:Bored, 5: Significantly bored)  
    \end{itemize}
    \bigskip

    \item \textbf{Overall Quality:}
    \begin{itemize}
        \item How would you rate your experience according to your expectations? (5-point Likert scale, 1: Significantly worse than expected, 2: Worse than expected, 3: Same as expected, 4: Better than expected, 5: Significantly better than expected)
    \end{itemize}
    
\end{enumerate}
\end{document}